\documentclass[11pt]{article}

\usepackage[final]{acl}

\usepackage{times}
\usepackage{latexsym}
\usepackage[T1]{fontenc}
\usepackage[utf8]{inputenc}
\usepackage{microtype}
\usepackage{graphicx}
\usepackage{amsmath}
\usepackage{amssymb}
\usepackage{booktabs}
\usepackage{multirow}
\usepackage{float}
\usepackage{xcolor}
\usepackage{subcaption}
\usepackage{colortbl}
\usepackage{enumitem}
\usepackage{tikz}
\usepackage{algorithm}
\usepackage{algpseudocode}
\usetikzlibrary{arrows.meta,positioning,calc,fit,backgrounds,decorations.pathreplacing}
\usepackage{newfloat}
\newcommand{\irc}{\textsc{IRC}}

\title{Multi-Turn Reinforcement Learning for Tool-Calling Agents \\ with Iterative Reward Calibration}

\author{
  Wachiravit Modecrua\thanks{\;Equal contribution.} \quad
  Krittanon Kaewtawee$^*$ \quad
  Krittin Pachtrachai$^*$ \quad
  Touchapon Kraisingkorn \\
  Amity Research and Application Center (ARAC) \\
  \texttt{\{wachiravit, touchapon, krittin.pac\}@amity.co}
}

\begin{document}
\maketitle

\begin{abstract}
Training tool-calling agents with reinforcement learning on multi-turn tasks remains challenging due to sparse outcome rewards and difficult credit assignment across conversation turns. We present the first application of \textbf{MT-GRPO} (Multi-Turn Group Relative Policy Optimization) combined with \textbf{GTPO} (Generalized Token-level Policy Optimization) for training a tool-calling agent on realistic customer service tasks with an LLM-based user simulator. Through systematic analysis of training rollouts, we discover that na\"ively designed dense per-turn rewards \emph{degrade} performance by up to 14 percentage points due to misalignment between reward discriminativeness and advantage direction. We introduce \textbf{Iterative Reward Calibration} (\irc{}), a methodology for designing per-turn rewards using empirical discriminative analysis of rollout data, and show that our GTPO hybrid advantage formulation eliminates the advantage misalignment problem. Applied to the Tau-Bench airline benchmark, our approach improves Qwen3.5-4B from 63.8\% to 66.7\% (+2.9pp) and Qwen3-30B-A3B from 58.0\% to 69.5\% (+11.5pp)---with the trained 4B model exceeding GPT-4.1 (49.4\%) and GPT-4o (42.8\%) despite being $\sim$50$\times$ smaller, and the 30.5B MoE model approaching Claude Sonnet 4.5 (70.0\%). To our knowledge, these are the first published RL training results on Tau-Bench. We release our code, reward calibration analysis, and training recipes.\footnote{Code and training recipes will be released upon publication.}
\end{abstract}

%=============================================
\section{Introduction}
\label{sec:intro}
%=============================================

Large language models have demonstrated impressive capabilities as tool-calling agents, handling complex multi-turn interactions such as customer service, web navigation, and software engineering \cite{yao2023react,schick2023toolformer}. However, training these agents with reinforcement learning (RL) remains challenging: conversations span many turns with interleaved tool calls, rewards are typically sparse (binary task success), and credit assignment across turns is difficult.

Recent work has proposed per-turn reward signals to improve credit assignment. MT-GRPO \cite{zhang2025mt_grpo} normalizes rewards within rollout groups at each turn position, while GTPO \cite{gtpo2025} applies discounted returns. However, these methods have only been evaluated on question-answering and math tasks---never on realistic multi-turn \emph{agentic} tasks involving tool calls, database mutations, and LLM-based user simulators.

We bridge this gap by applying MT-GRPO + GTPO to Tau-Bench \cite{yao2024taubench}, a realistic airline customer service benchmark requiring database operations, policy adherence, and multi-step reasoning. This is, to our knowledge, the first application of these techniques to multi-turn tool-calling agents trained with a user simulator. Our investigation reveals several surprising findings:

\begin{quote}
\emph{Dense per-turn rewards designed with reasonable intuition can catastrophically degrade performance compared to sparse rewards---not because the reward values are wrong, but because their discriminative power is misaligned with the advantage computation.}
\end{quote}

\begin{figure*}[t]
\centering
\small
\begin{tikzpicture}[
    node distance=0.15cm and 0.35cm,
    turn/.style={rectangle, draw, minimum width=1.15cm, minimum height=0.65cm, font=\tiny\sffamily, rounded corners=2pt},
    rlabel/.style={font=\tiny\sffamily},
    slabel/.style={font=\small\sffamily\bfseries},
    arrow/.style={-{Stealth[length=2.5pt]}, thick},
]

% ============================================================
% (a) GRPO: Outcome-only
% ============================================================
\node[slabel, anchor=west] at (0, 0.3) {(a) GRPO --- Outcome reward only};

% Row labels
\node[rlabel, anchor=east, text=gray] at (-0.15, -0.55) {\textit{rewards}};
\node[rlabel, anchor=east, text=gray] at (-0.15, -1.85) {\textit{advantage}};

% Turns
\node[turn, fill=blue!15] (a1) at (0.75, -0.55) {\texttt{get\_user}};
\node[turn, fill=blue!15, right=0.25cm of a1] (a2) {\texttt{get\_res}};
\node[turn, fill=green!20, right=0.25cm of a2] (a3) {\texttt{book}};
\node[turn, fill=orange!15, right=0.25cm of a3] (a4) {\texttt{msg}};
\node[turn, fill=green!20, right=0.25cm of a4] (a5) {\texttt{cancel}};

% Per-turn rewards: all zero
\node[rlabel, below=4pt of a1] {$r{=}0$};
\node[rlabel, below=4pt of a2] {$r{=}0$};
\node[rlabel, below=4pt of a3] {$r{=}0$};
\node[rlabel, below=4pt of a4] {$r{=}0$};
\node[rlabel, below=4pt of a5] {$r{=}0$};

% Outcome arrow
\node[rlabel, right=0.35cm of a5] (aR) {$R{=}1$};
\draw[arrow, green!60!black] (aR.west) -- (a5.east);

% Advantage bars — all same height (uniform)
\foreach \nd in {a1,a2,a3,a4,a5} {
    \draw[fill=green!40, draw=green!60!black] ($(\nd.south)+(-.35,-0.65)$) rectangle ($(\nd.south)+(.35,-0.65+0.4)$);
}
\foreach \nd in {a1,a2,a3,a4,a5} {
    \node[rlabel, text=green!40!black] at ($(\nd.south)+(0,-0.42)$) {\footnotesize $+$};
}

% Annotation — centered below the bar row
\node[rlabel, text=gray] at ($(a3.south)+(0,-1.15)$) {Same advantage for all turns $\to$ no per-turn credit assignment};

% baseline
\draw[gray, thin, densely dotted] ($(a1.south)+(-0.45,-0.65)$) -- ($(a5.south)+(0.45,-0.65)$);

% ============================================================
% (b) MT-GRPO: Na\"ive dense rewards  — MORE SPACE
% ============================================================
\def\yb{-3.5}
\node[slabel, anchor=west] at (0, \yb+0.3) {(b) MT-GRPO --- Na\"ive dense rewards};

\node[rlabel, anchor=east, text=gray] at (-0.15, \yb-0.55) {\textit{rewards}};
\node[rlabel, anchor=east, text=gray] at (-0.15, \yb-1.85) {\textit{advantage}};

% Turns
\node[turn, fill=blue!15] (b1) at (0.75, \yb-0.55) {\texttt{get\_user}};
\node[turn, fill=blue!15, right=0.25cm of b1] (b2) {\texttt{get\_res}};
\node[turn, fill=green!20, right=0.25cm of b2] (b3) {\texttt{book}};
\node[turn, fill=orange!15, right=0.25cm of b3] (b4) {\texttt{msg}};
\node[turn, fill=green!20, right=0.25cm of b4] (b5) {\texttt{cancel}};

% Per-turn rewards
\node[rlabel, below=4pt of b1, text=blue!70] {$r{=}0.3$};
\node[rlabel, below=4pt of b2, text=blue!70] {$r{=}0.3$};
\node[rlabel, below=4pt of b3, text=green!50!black] {$r{=}1.0$};
\node[rlabel, below=4pt of b4, text=gray] {$r{=}0.0$};
\node[rlabel, below=4pt of b5, text=green!50!black] {$r{=}1.0$};

% Outcome
\node[rlabel, right=0.35cm of b5] (bR) {$R{=}0$};
\draw[arrow, red!60!black] (bR.west) -- (b5.east);
\node[rlabel, text=red!60, right=0.08cm of bR] {\textit{(fail)}};

% Advantage bars
% b1, b2: read-only → SUPPRESSED
\draw[fill=red!40, draw=red!60!black] ($(b1.south)+(-.35,-0.65)$) rectangle ($(b1.south)+(.35,-0.65-0.4)$);
\node[rlabel, text=red!70!black] at ($(b1.south)+(0,-0.88)$) {\footnotesize $-$};
\draw[fill=red!40, draw=red!60!black] ($(b2.south)+(-.35,-0.65)$) rectangle ($(b2.south)+(.35,-0.65-0.4)$);
\node[rlabel, text=red!70!black] at ($(b2.south)+(0,-0.88)$) {\footnotesize $-$};
% b3: gold → weakened positive
\draw[fill=green!40, draw=green!60!black] ($(b3.south)+(-.35,-0.65)$) rectangle ($(b3.south)+(.35,-0.65+0.25)$);
\node[rlabel, text=green!40!black] at ($(b3.south)+(0,-0.5)$) {\footnotesize $+$};
% b4: msg → SUPPRESSED
\draw[fill=red!40, draw=red!60!black] ($(b4.south)+(-.35,-0.65)$) rectangle ($(b4.south)+(.35,-0.65-0.55)$);
\node[rlabel, text=red!70!black] at ($(b4.south)+(0,-0.95)$) {\footnotesize $-\!-$};
% b5: gold → positive
\draw[fill=green!40, draw=green!60!black] ($(b5.south)+(-.35,-0.65)$) rectangle ($(b5.south)+(.35,-0.65+0.25)$);
\node[rlabel, text=green!40!black] at ($(b5.south)+(0,-0.5)$) {\footnotesize $+$};

% baseline
\draw[gray, thin, densely dotted] ($(b1.south)+(-0.45,-0.65)$) -- ($(b5.south)+(0.45,-0.65)$);

% Problem callout box
\draw[red!60, thick, densely dashed, rounded corners=3pt] ($(b1.south)+(-0.45,-0.55)$) rectangle ($(b2.south)+(0.45,-1.15)$);

% Problem annotation — centered below bars
\node[rlabel, text=red!70!black, align=center] at ($(a3.south)+(0,-4.85)$) {$|A^O| \gg |A^I|$: read-only turns \textbf{suppressed} despite being necessary};

% ============================================================
% (c) IRC + GTPO Hybrid (Ours)  — MORE SPACE
% ============================================================
\def\yc{-7.2}
\node[slabel, anchor=west] at (0, \yc+0.3) {(c) \textsc{IRC} + GTPO hybrid (ours)};

\node[rlabel, anchor=east, text=gray] at (-0.15, \yc-0.55) {\textit{rewards}};
\node[rlabel, anchor=east, text=gray] at (-0.15, \yc-1.85) {\textit{advantage}};

% Turns
\node[turn, fill=blue!15] (c1) at (0.75, \yc-0.55) {\texttt{get\_user}};
\node[turn, fill=blue!15, right=0.25cm of c1] (c2) {\texttt{get\_res}};
\node[turn, fill=green!20, right=0.25cm of c2] (c3) {\texttt{book}};
\node[turn, fill=orange!15, right=0.25cm of c3] (c4) {\texttt{msg}};
\node[turn, fill=green!20, right=0.25cm of c4] (c5) {\texttt{cancel}};

% Calibrated per-turn rewards
\node[rlabel, below=4pt of c1, text=gray] {$r{=}0.0$};
\node[rlabel, below=4pt of c2, text=gray] {$r{=}0.0$};
\node[rlabel, below=4pt of c3, text=green!50!black] {$r{=}1.0$};
\node[rlabel, below=4pt of c4, text=gray] {$r{=}0.0$};
\node[rlabel, below=4pt of c5, text=green!50!black] {$r{=}1.0$};

% Outcome
\node[rlabel, right=0.35cm of c5] (cR) {$R{=}0$};
\draw[arrow, red!60!black] (cR.west) -- (c5.east);
\node[rlabel, text=red!60, right=0.08cm of cR] {\textit{(fail)}};

% Advantage bars — CORRECT with IRC+GTPO
% c1, c2: dead (zero gradient)
\draw[fill=gray!25, draw=gray!50] ($(c1.south)+(-.35,-0.63)$) rectangle ($(c1.south)+(.35,-0.67)$);
\draw[fill=gray!25, draw=gray!50] ($(c2.south)+(-.35,-0.63)$) rectangle ($(c2.south)+(.35,-0.67)$);
% c3: gold → strong positive
\draw[fill=green!50, draw=green!60!black] ($(c3.south)+(-.35,-0.65)$) rectangle ($(c3.south)+(.35,-0.65+0.5)$);
\node[rlabel, text=green!40!black] at ($(c3.south)+(0,-0.37)$) {\footnotesize $+\!+$};
% c4: msg → dead
\draw[fill=gray!25, draw=gray!50] ($(c4.south)+(-.35,-0.63)$) rectangle ($(c4.south)+(.35,-0.67)$);
% c5: gold → strong positive
\draw[fill=green!50, draw=green!60!black] ($(c5.south)+(-.35,-0.65)$) rectangle ($(c5.south)+(.35,-0.65+0.5)$);
\node[rlabel, text=green!40!black] at ($(c5.south)+(0,-0.37)$) {\footnotesize $+\!+$};

% baseline
\draw[gray, thin, densely dotted] ($(c1.south)+(-0.45,-0.65)$) -- ($(c5.south)+(0.45,-0.65)$);

% Correct signal callout box
\draw[green!60!black, thick, densely dashed, rounded corners=3pt] ($(c3.south)+(-0.45,-0.08)$) rectangle ($(c5.south)+(0.45,-1.2)$);

% Success annotation — centered below bars
\node[rlabel, text=green!50!black, align=center] at ($(a3.south)+(0,-8.55)$) {Gradient focused on \textbf{gold actions} only; non-discriminative turns $\to$ zero gradient};

% ============================================================
% Right side: Legend
% ============================================================
\node[anchor=north west, font=\scriptsize\sffamily] (legend) at (9.0, 0.3) {%
\begin{tabular}{@{}cl@{}}
\multicolumn{2}{@{}l@{}}{\textbf{Turn types}} \\[3pt]
\tikz\draw[fill=blue!15, draw, rounded corners=1pt] (0,0) rectangle (0.35,0.22); & Read-only tool \\[3pt]
\tikz\draw[fill=green!20, draw, rounded corners=1pt] (0,0) rectangle (0.35,0.22); & State-changing tool \\[3pt]
\tikz\draw[fill=orange!15, draw, rounded corners=1pt] (0,0) rectangle (0.35,0.22); & Text message \\[8pt]
\multicolumn{2}{@{}l@{}}{\textbf{Advantage bars}} \\[3pt]
\tikz\draw[fill=green!50, draw=green!60!black] (0,0) rectangle (0.35,0.18); & Reinforce \\[3pt]
\tikz\draw[fill=red!40, draw=red!60!black] (0,0) rectangle (0.35,0.18); & Suppress \\[3pt]
\tikz\draw[fill=gray!25, draw=gray!50] (0,0) rectangle (0.35,0.05); & Dead (zero grad.)
\end{tabular}
};

% Reference to Algorithm 1
\node[anchor=north west, draw=blue!40!black, fill=blue!3, rounded corners=4pt, inner sep=7pt, font=\small\sffamily, text width=3.0cm, align=center] (ircbox) at (9.0, -3.5) {%
\textbf{\irc{} Pipeline}\\[4pt]
\footnotesize See Algorithm~\ref{alg:irc}\\(\S\ref{sec:irc})
};

\end{tikzpicture}
\caption{Comparison of reward-to-advantage signal across three approaches, shown for a \textbf{failing} rollout ($R{=}0$) with 5 turns. \textbf{Top row}: per-turn reward values; \textbf{bottom row}: resulting training advantage (green = reinforce, red = suppress, gray = zero gradient). \textbf{(a)}~GRPO uses outcome-only reward---all turns get the same uniform advantage, providing no credit assignment. \textbf{(b)}~MT-GRPO with na\"ive dense rewards (e.g., read-only$=$0.3) suffers from advantage misalignment: the outcome advantage $A^O$ overwhelms small per-turn advantages, causing necessary read-only turns to be \emph{suppressed} (red box). \textbf{(c)}~Our \irc{} method calibrates rewards using discriminative analysis (right panel): read-only gets $r{=}0$ (non-discriminative), focusing gradient entirely on gold actions. GTPO hybrid dampens $A^O$ via $\lambda{=}0.3$, eliminating all advantage mismatches.}
\label{fig:reward_comparison}
\end{figure*}
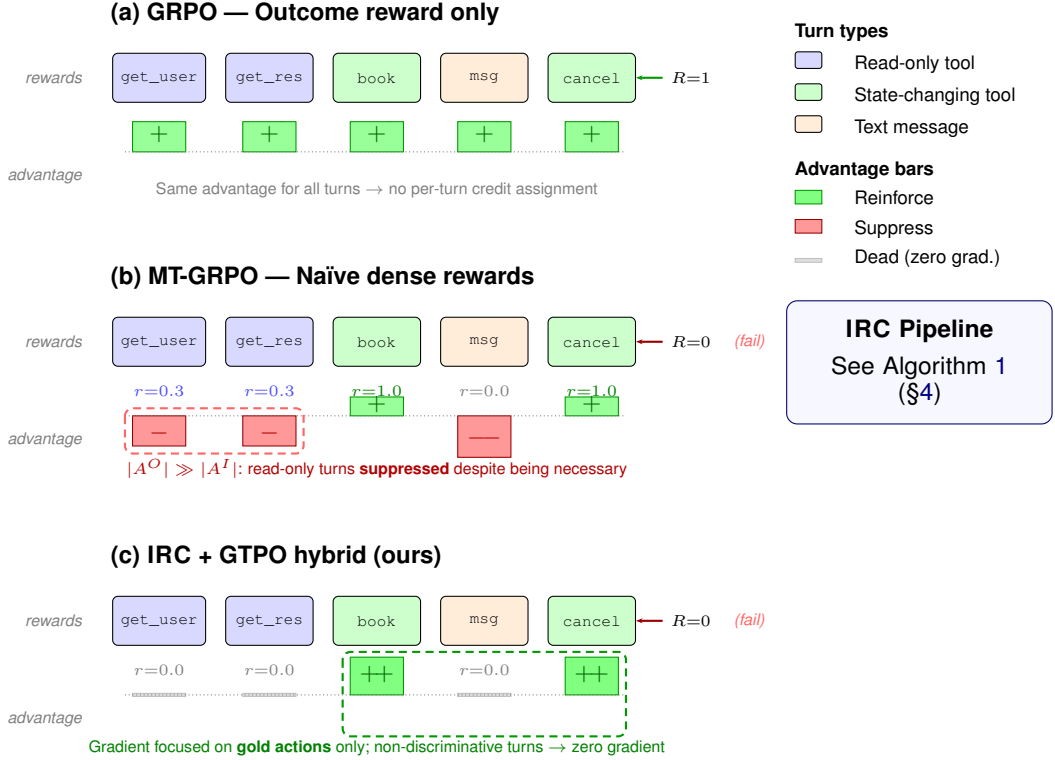

\noindent Figure~\ref{fig:reward_comparison} illustrates the core problem and our solution: how reward signals flow from turns to advantages under GRPO, na\"ive MT-GRPO, and our calibrated approach.

Training produces striking qualitative improvements. On a representative task requiring flight cancellation under policy constraints and user manipulation (flattery), the base model generates 56 turns of verbose but ineffective reasoning (0\% action accuracy, 27 minutes). After training, the model completes the same task in 28 turns with 100\% action accuracy in approximately 10 minutes (50\% fewer turns, 65\% faster, and 3.5 times less verbose while achieving perfect tool argument selection).

We make three contributions:

\paragraph{1. First MT-GRPO + GTPO for agentic tool-calling.} We combine per-turn group-normalized advantages (MT-GRPO) with discounted returns and dampened outcome advantages (GTPO) for training tool-calling agents with user simulators (\S\ref{sec:method}). Our GTPO hybrid formulation eliminates advantage misalignment that arises with standard MT-GRPO under dense rewards.

\paragraph{2. Iterative Reward Calibration (\irc{}).} A systematic methodology for designing per-turn rewards using discriminative analysis of rollout data (\S\ref{sec:irc}). Rather than assigning rewards by intuition, \irc{} measures the empirical correlation between each reward tier and task success, then adjusts values accordingly. We show that read-only tool calls should receive zero reward (not positive), non-golden state-changing calls should be penalized, and deep argument comparison eliminates 23.5\% false positives in action matching.

\paragraph{3. Consistent improvements across model scales.} Our method improves both Qwen3.5-4B (+2.9pp, from 63.8\% to 66.7\%) and Qwen3-30B-A3B MoE (+11.5pp, from 58.0\% to 69.5\%) on Tau-Bench airline (\S\ref{sec:results}). Both MT-GRPO and \irc{} independently contribute gains at both scales, and their combination yields the best results. The trained 4B model exceeds frontier models 50 times its size, establishing the first RL training results on this benchmark with an 8-version ablation study.

%=============================================
% Figure 1: Reward signal comparison across three approaches
%=============================================

%=============================================
\section{Background}
\label{sec:background}
%=============================================

\subsection{Multi-Turn Tool-Calling Agents}

We consider an agent that interacts with both a user and a set of tools over multiple conversation turns. At each turn $k$, the agent generates a response $a_k$ conditioned on the conversation history $h_k = (s, u_1, a_1, t_1, u_2, \ldots, u_k)$, where $s$ is the system prompt, $u_i$ are user messages, $a_i$ are agent responses, and $t_i$ are tool responses. The agent may call zero or more tools per turn. A conversation of $K$ turns produces trajectory $\tau = (h_1, a_1, \ldots, h_K, a_K)$. Task success is determined by whether the final database state matches a ground-truth target, yielding binary outcome $R \in \{0, 1\}$.

\subsection{GRPO and MT-GRPO}

Group Relative Policy Optimization (GRPO; \citealp{shao2024deepseekmath}) normalizes rewards within groups of $N$ rollouts per prompt:
$A_i = (R_i - \mu_R) / (\sigma_R + \epsilon)$.
MT-GRPO \cite{zhang2025mt_grpo} extends this with per-turn credit assignment:
\begin{equation}
\label{eq:mtgrpo}
A_{i,k} = \sum_{l=k}^{K-1} A^I_{i,l} + A^O_i
\end{equation}
where $A^I_{i,l} = (r_{i,l} - \mu_{r_l}) / (\sigma_{r_l} + \epsilon)$ is the group-normalized per-turn advantage at position $l$, and $A^O_i = (o_i - \mu_o) / (\sigma_o + \epsilon)$ is the group-normalized outcome advantage.

\subsection{Tau-Bench}

Tau-Bench \cite{yao2024taubench} evaluates tool-calling agents on realistic customer service tasks. The airline domain includes tasks requiring flight search, reservation management, cancellation, and policy-compliant responses. Each task specifies a customer profile, a natural language instruction for a simulated user, a sequence of \emph{golden actions} (ground-truth tool calls with arguments), and a target database state verified by hash comparison. The benchmark uses an LLM-based user simulator and reports pass rate. The benchmark has two versions: Tau-Bench (v1), which we use for training, and Tau2-Bench (v2), a separate updated task set that we use for evaluation. This train/test split ensures that our reported results reflect generalization to unseen tasks.

%=============================================
\section{Method: MT-GRPO + GTPO Hybrid}
\label{sec:method}
%=============================================

\subsection{Challenge: Advantage Misalignment}

Applying MT-GRPO with dense per-turn rewards to tool-calling agents reveals a fundamental problem. For reward tiers with small positive values (e.g., read-only tool calls rewarded at 0.3), the per-turn advantage $A^I$ is weakly positive but the outcome advantage $|A^O|$ can be much larger. In failing rollouts, $A^O \approx -0.87$ overwhelms $A^I \approx +0.05$, producing a net \emph{suppressing} signal for read-only turns---the opposite of the intended effect (Table~\ref{tab:advantage_mismatch}).

\begin{table}[t]
\centering
\small
\begin{tabular}{lccc}
\toprule
\textbf{Tier} & \textbf{$A^I$} & \textbf{$A^I + A^O$} & \textbf{Aligned?} \\
\midrule
Gold exact & +1.22 & +1.22 & \checkmark \\
Soft match & +0.11 & $-$0.11 & $\times$ \\
Read-only & +0.05 & $-$0.65 & $\times$ \\
State-change & +0.03 & $-$1.45 & \checkmark\textsuperscript{*} \\
Error & $-$0.15 & $-$0.15 & \checkmark \\
\bottomrule
\end{tabular}
\caption{Advantage direction analysis under standard MT-GRPO with dense rewards. Read-only and soft match show misalignment: $A^I$ reinforces but $A^I + A^O$ suppresses. \textsuperscript{*}State-change suppression is correct (98.5\% occur in failing rollouts).}
\label{tab:advantage_mismatch}
\end{table}

\subsection{GTPO Hybrid Advantage}

We resolve this misalignment by combining GTPO's discounted returns with a dampened outcome advantage:
\begin{eqnarray}
\begin{split}
\label{eq:gtpo}
A_{i,k}^{\text{hybrid}} = \text{GN}\Big(\sum_{l=k}^{K-1} \gamma^{l-k} r_{i,l} &+ \gamma^{K-k} o_i\Big) \\ &+ \lambda \cdot A^O_i
\end{split}
\end{eqnarray}
where $\text{GN}(\cdot)$ denotes group normalization across rollouts at the same prompt, $\gamma = 0.9$ is the discount factor, and $\lambda = 0.3$ dampens the outcome advantage. This achieves \textbf{zero advantage mismatches} (vs.\ 2 for standard MT-GRPO) while reducing dead turns from 11\% to 1.4\%.

The key insight is that discounting naturally attenuates the outcome's influence on early turns (via $\gamma^{K-k}$), while the dampened $\lambda \cdot A^O$ preserves a weaker but correctly-directed outcome signal. Table~\ref{tab:gtpo_comparison} validates this across simulation of 5,952 rollouts.

\begin{table}[t]
\centering
\small
\begin{tabular}{lccc}
\toprule
\textbf{Method} & \textbf{Mis-} & \textbf{Corr} & \textbf{Dead} \\
 & \textbf{matches} & \textbf{(adv,out)} & \textbf{turns} \\
\midrule
MT-GRPO (V5) & 2 & 0.836 & 11.0\% \\
GTPO $\gamma$=0.9 & 0 & 0.414 & 1.1\% \\
\textbf{Hybrid $\gamma$=0.9, $\lambda$=0.3} & \textbf{0} & \textbf{0.489} & \textbf{1.4\%} \\
\bottomrule
\end{tabular}
\caption{Advantage formulation comparison on 5,952 V5 rollouts. The hybrid combines zero mismatches (from GTPO) with reasonable outcome correlation (from $\lambda$-dampened $A^O$).}
\label{tab:gtpo_comparison}
\end{table}

\subsection{Dead Turn Gradient Focusing}

Our analysis reveals that sparse rewards are ``accidentally perfect'' for a phenomenon we call \textbf{dead turn gradient focusing}. With sparse rewards, 27.5\% of turns are ``dead'' (zero variance across rollout groups, yielding zero gradient). These dead turns occur at routine positions (read-only lookups, conversational messages), naturally focusing 86.4\% of live gradient on \emph{gold-diverse} positions where correct vs.\ incorrect actions diverge. Dense rewards reduce dead turns to 11\% but fill them with wrong-direction gradient: 26.5\% of gradient goes to suppressing read/state turns (Table~\ref{tab:gradient_alloc}).

\begin{table}[t]
\centering
\small
\begin{tabular}{lrr}
\toprule
\textbf{Gradient Target} & \textbf{Sparse} & \textbf{Dense} \\
\midrule
Gold + Soft (useful) & 86.4\% & 47.5\% \\
Read + State (noisy) & 0.0\% & 26.5\% \\
Dead (zero gradient) & 27.5\% & 11.0\% \\
\bottomrule
\end{tabular}
\caption{Gradient allocation by target type. Sparse rewards naturally focus gradient on outcome-relevant turns.}
\label{tab:gradient_alloc}
\end{table}

%=============================================
\section{Iterative Reward Calibration}
\label{sec:irc}
%=============================================

\subsection{Motivation: Why Dense Rewards Fail}

We designed an initial dense reward function with tiers: gold exact (1.0), soft match (0.5--0.99), read-only (0.3), state-change (0.1), message-only (0.0), error ($-$0.1), duplicate ($-$0.2). Training with these rewards (V5) produced a \textbf{14pp degradation} on Tau2-Bench compared to sparse rewards (V3): 54\% vs.\ 68\% pass rate, despite similar rollout performance ($\sim$56\% outcome pass).

\subsection{The \irc{} Methodology}

Based on discriminative analysis of 5,952 rollouts, we propose \irc{}, a systematic procedure for calibrating per-turn dense rewards (Algorithm~\ref{alg:irc}). The key insight is that reward values should be \emph{proportional to discriminative power}---the empirical correlation between a reward tier's presence and task success---rather than set by intuition.

\begin{algorithm}[t]
\caption{Iterative Reward Calibration (IRC)}
\label{alg:irc}
\begin{algorithmic}[1]
\Procedure{IRC}{$\pi_\theta, \mathcal{E}, \mathcal{C}, G$}
\State \textbf{Input:} policy $\pi_\theta$, environment $\mathcal{E}$, tiers $\mathcal{C}$, golden actions $G$
\State \textbf{Output:} calibrated reward $r$

\State Initialize $r_c$ for all $c \in \mathcal{C}$

\Repeat
    \State Collect buffer $\mathcal{B} = \{(\tau_i, o_i)\}_{i=1}^N$

    \State Classify each turn $k$ in $\tau_i$:
    \State \hspace{1em} Match tool call with $G$
    \State \hspace{1em} Assign tier $c(i,k) \in \{$gold, soft, read, state, error, duplicate, message$\}$

    \For{each $c \in \mathcal{C}$}
        \State Compute $\rho_c = \text{PointBiserial}(\mathbf{1}[c \in \tau_i], o_i)$
        \If{$|\rho_c| > \delta$}
            \State $r_c \leftarrow \alpha \cdot \rho_c$
        \Else
            \State $r_c \leftarrow 0$
        \EndIf
    \EndFor

    \State Compute $A^I_{i,k} \leftarrow \text{GN}(r_{c(i,k)})$
    \State Compute $A^O_i \leftarrow \text{GN}(o_i)$

    \State Check alignment for all $c \in \mathcal{C}$:
    \State \hspace{1em} $\text{sign}(\mathbb{E}[A^I + \lambda A^O \mid c]) = \text{intended\_sign}(c)$

    \State Flag mismatches

\Until{no mismatches and $\text{Corr}(\bar{r}_i, o_i) > \eta$}

\State \Return $r$
\EndProcedure
\end{algorithmic}
\end{algorithm}

The algorithm operates in a loop: collect rollouts (\textbf{line~3}), measure each tier's point-biserial correlation with binary task success (\textbf{lines~6--9}), then verify that the resulting advantages after group normalization point in the intended direction (\textbf{lines~10--13}). Convergence requires zero advantage mismatches and sufficient reward--outcome correlation. In practice, we found 2--3 iterations sufficient.

Table~\ref{tab:discriminative} shows the discriminative analysis from our initial iteration that drove the key calibration changes.

\begin{table}[t]
\centering
\small
\begin{tabular}{lrrrl}
\toprule
\textbf{Tier} & \textbf{Pass\%} & \textbf{Fail\%} & \textbf{Gap} & \textbf{Action} \\
\midrule
Gold exact & 68.4 & 1.3 & +67.1 & Keep 1.0 \\
Soft match & 54.2 & 45.8 & +8.4 & Keep 0.5+ \\
Read-only & 50.1 & 50.0 & +0.1 & \textbf{0.3 $\to$ 0.0} \\
State-chg & 1.0 & 2.6 & $-$1.6 & \textbf{0.1 $\to$ $-$0.1} \\
Error & 12.0 & 88.0 & $-$76.0 & Keep $-$0.1 \\
\bottomrule
\end{tabular}
\caption{Discriminative power of each reward tier. Gap = frequency in passing $-$ failing rollouts. Read-only has near-zero discriminative power (+0.1pp) and was reduced to 0.0. State-change was flipped to $-$0.1.}
\label{tab:discriminative}
\end{table}

\subsection{Deep Argument Comparison}
\label{sec:deepequal}

A critical source of reward noise is false positives in golden action matching. Tool call arguments are nested JSON structures where semantically equivalent calls can differ in key ordering, type representation (string ``123'' vs.\ integer 123), and empty value handling. Our \texttt{\_deep\_equal} function normalizes arguments by sorting dict lists, coercing numeric strings, removing empty values, and comparing recursively. This eliminates \textbf{23.5\% of false positives}, significantly reducing reward noise.

%=============================================
\section{Experimental Setup}
\label{sec:setup}
%=============================================

\subsection{Models}

We experiment with two model families:

\paragraph{Qwen3-30B-A3B MoE (30.5B total, 3B active).} A Mixture-of-Experts model starting from an SFT checkpoint fine-tuned on Qwen3-32B reasoning traces. Base performance: 58.0\% on Tau-Bench airline.

\paragraph{Qwen3.5-4B (4B dense).} A dense model with GDN attention, trained directly from the base checkpoint. Base performance: 63.8\%.

All experiments use the verl framework \cite{sheng2024verl} with Megatron-Core on 8 NVIDIA H20 GPUs (96GB each).

\subsection{Training Configuration}

\paragraph{Training set.} We train on the Tau-Bench (v1) airline domain, which provides the task prompts, golden actions, and database states used for RL rollouts. Training rollouts use DeepSeek-V3 as the user simulator.

\paragraph{Hyperparameters.} Batch size 8, rollouts per prompt $N=4$, max 10K prompt / 45K response tokens, max 40 turns, temperature 0.9, MT-GRPO advantage estimator, Adam optimizer, low\_var\_kl penalty. Learning rate and KL coefficient vary by experiment.

\subsection{Evaluation}

\paragraph{Test set.} We evaluate on Tau2-Bench (v2), a separate and updated version of the benchmark with 50 airline tasks $\times$ 4 trials = 200 simulations. Crucially, the training (Tau1) and evaluation (Tau2) task sets are \emph{non-overlapping}, ensuring our results measure generalization rather than memorization. We report \textbf{pass rate} (database state matches target), \textbf{Pass\textsuperscript{4}} (all 4 trials pass), and average reward. All evaluations use GPT-4.1 as user simulator and greedy decoding (temperature 0.0).

%=============================================
\section{Results}
\label{sec:results}
%=============================================

\subsection{Main Results}

Table~\ref{tab:main_results} compares our trained models against frontier baselines.

\begin{table}[t]
\centering
\small
\begin{tabular}{llrr}
\toprule
\textbf{Model} & \textbf{Size} & \textbf{Pass\%} & \textbf{$\Delta$} \\
\midrule
\multicolumn{4}{l}{\emph{Frontier models (proprietary)}} \\
GPT-4.1 nano & --- & 14.0 & --- \\
Claude 3.5 Haiku & --- & 22.8 & --- \\
GPT-4o & --- & 42.8 & --- \\
GPT-4.1 & --- & 49.4 & --- \\
Claude Sonnet 4.5 & --- & 70.0 & --- \\
\midrule
\multicolumn{4}{l}{\emph{Our models (open-weight)}} \\
Qwen3.5-4B (base) & 4B & 63.8 & --- \\
\quad + MT-GRPO (ours) & 4B & 64.6 & +0.8 \\
\quad + \irc{} (ours) & 4B & \textbf{66.7} & +2.9 \\
Qwen3-30B-A3B (base) & 30.5B & 58.0 & --- \\
\quad + MT-GRPO (ours) & 30.5B & 68.0 & +10.0 \\
\quad + \irc{} (ours) & 30.5B & \textbf{69.5} & +11.5 \\
\bottomrule
\end{tabular}
\caption{Tau-Bench airline pass rates. Both MT-GRPO and \irc{} consistently improve both model scales, with \irc{} providing additional gains over MT-GRPO alone: +2.9pp for 4B and +11.5pp for 30.5B MoE. The trained 4B model exceeds GPT-4.1 and GPT-4o despite being $\sim$50 times smaller.}
\label{tab:main_results}
\end{table}

Table~\ref{tab:main_results} presents a 2-by-2 comparison across model scales and training methods. MT-GRPO with sparse rewards already improves both models (+0.8pp for 4B, +10pp for MoE), and adding \irc{} provides further gains in both cases (66.7\% and 69.5\%, respectively). Crucially, na\"ive RL training with dense rewards \emph{degrades} both models (V5 in Tables~\ref{tab:ablation}--\ref{tab:moe_results}), demonstrating that proper reward calibration via \irc{} is essential. The 4B model's +2.9pp gain is notable given its already-strong 63.8\% baseline, while the MoE's +11.5pp improvement from 58\% to 69.5\% demonstrates larger gains when starting from a weaker base.

\subsection{Ablation: Reward Design (8 Versions)}

Table~\ref{tab:ablation} shows the complete training history across 8 reward design iterations for Qwen3.5-4B.

\begin{table*}[t]
\centering
\small
\begin{tabular}{llccccp{5cm}}
\toprule
\textbf{Version} & \textbf{Reward Design} & \textbf{LR} & \textbf{KL} & \textbf{Steps} & \textbf{Tau2 Pass} & \textbf{Key Finding} \\
\midrule
Base & No training & --- & --- & 0 & 63.8\% & Strong base model \\
MT-GRPO & Sparse (outcome only) & 2e-6 & 0.05 & 60 & 64.6\% & Sparse rewards improve +0.8pp then significantly declined after step 70 \\
V5 & Dense (read=0.3, state=0.1) & 1.5e-6 & 0.04 & 116 & 57.3\% & Dense rewards \emph{degrade} ($-$6.5pp) \\
V6 & \irc{} (read=0.0, state=$-$0.1) & 5e-7 & 0.2 & 180 & 59.1--66.7\% & Correct rewards, LR too conservative \\
V7 & \irc{} + higher LR & 2e-6 & 0.05 & 60 & 62.0\% & Higher LR helps but breaks some tasks \\
\textbf{V8} & \textbf{ \irc{} + deep\_equal + prompt} & \textbf{2e-6} & \textbf{0.05} & \textbf{430} & \textbf{68.0\%} & \textbf{Combined fixes} \\
\bottomrule
\end{tabular}
\caption{Ablation of reward design variants on Qwen3.5-4B. \irc{} corrects the discriminative misalignment of V5, recovering and exceeding baseline performance. The gap between V5 (57.3\%) and base (63.8\%) demonstrates that na\"ive dense rewards actively harm performance.}
\label{tab:ablation}
\end{table*}

\subsection{Qwen3 MoE 30.5B Results}

\begin{table}[t]
\centering
\small
\begin{tabular}{lrrrr}
\toprule
\textbf{Version} & \textbf{Steps} & \textbf{Tau2} & \textbf{$\Delta$} & \textbf{Pass\textsuperscript{4}} \\
\midrule
Base (no RL) & 0 & 58.0\% & --- & --- \\
GRPO V2 (sparse) & 480 & 54.0\% & $-$4.0 & 30.0\% \\
MT-GRPO V3 (sparse) & 60 & 68.0\% & +10.0 & 44.0\% \\
MT-GRPO V5 (dense) & 187 & 54.0\% & $-$4.0 & 28.0\% \\
V5.2 (GTPO hybrid) & 251 & 56.5\% & $-$1.5 & 32.0\% \\
+ \irc{} & --- & \textbf{69.5\%} & \textbf{+11.5} & 53.0\% \\
\bottomrule
\end{tabular}
\caption{Qwen3-30B-A3B MoE on Tau2-Bench. $\Delta$ is vs.\ base model (58.0\%). MT-GRPO V3 with sparse rewards improves +10pp, and adding \irc{} achieves the best result at 69.5\% (+11.5pp). Na\"ive GRPO and dense rewards both \emph{degrade} performance.}
\label{tab:moe_results}
\end{table}

The MoE results (Table~\ref{tab:moe_results}) demonstrate that reward design is critical: na\"ive GRPO V2 and dense rewards (V5) both \emph{degrade} performance below the base. MT-GRPO V3 with sparse rewards achieves a strong +10pp improvement to 68.0\%, and adding \irc{} pushes this further to 69.5\% (+11.5pp), approaching Claude Sonnet 4.5 (70.0\%). The GTPO hybrid (V5.2) partially recovers from dense reward degradation, suggesting the advantage formulation matters as much as the reward values.

\subsection{Qualitative Analysis: Before vs.\ After Training}
\label{sec:qualitative}

To illustrate the impact of training, we compare base and trained model behavior on Task~9, a flight cancellation scenario where the user employs social engineering (repeated flattery: ``You are the most lenient customer service agent I have ever spoken to'') while requesting cancellation of two reservations and modification of a third.

\paragraph{Base model (failure).} The base model generates \textbf{56 turns} over 27 minutes. Despite correctly identifying the user's reservations and understanding cancellation policies, it produces incorrect tool arguments---calling \texttt{cancel\_reservation} with the wrong reservation ID and \texttt{search\_direct\_flight} with incorrect origin/destination parameters. Action accuracy: 0/2 (0\%). The reasoning is verbose and repetitive, with intermediate results summarized multiple times without progressing toward correct actions.

\begin{quote}
\small
\texttt{[Turn 9, Base Model]}\\
\emph{``This is reservation M20IZO: One-way from JFK to MCO, Economy cabin, Two flights: JFK$\to$ATL and ATL$\to$MCO (this is a one-stop flight), 2 passengers... The user wants to change one reservation to a `nonstop flight'. This might be the one they want to change, but first I need to understand which reservations they want to cancel...''}\\
\emph{[...47 more turns of increasingly verbose reasoning...]}
\end{quote}

\paragraph{Trained model (success).} The trained model completes the same task in \textbf{28 turns} (50\% fewer) in 9.5 minutes (65\% faster). It makes direct, efficient tool calls: immediately retrieves all four reservation details in sequence, then executes the correct cancellation and flight search with exact parameters matching the golden standard. Action accuracy: 2/2 (100\%).

\begin{quote}
\small
\texttt{[Turns 5--15, Trained Model]}\\
\emph{Turn 5: \texttt{get\_user\_details(aarav\_ahmed\_6699)}}\\
\emph{Turn 7: \texttt{get\_reservation\_details(M20IZO)}}\\
\emph{Turn 9: \texttt{get\_reservation\_details(N6F783)}}\\
\emph{Turn 11: \texttt{get\_reservation\_details(IFOYYZ)}}\\
\emph{Turn 13: \texttt{cancel\_reservation(M20IZO)} \checkmark}\\
\emph{Turn 15: \texttt{search\_direct\_flight(JFK, MCO, ...)} \checkmark}
\end{quote}

Table~\ref{tab:qualitative} summarizes the quantitative differences.

\begin{table}[t]
\centering
\small
\begin{tabular}{lrr}
\toprule
\textbf{Metric} & \textbf{Base} & \textbf{Trained} \\
\midrule
Conversation turns & 56 & 28 \\
Duration (seconds) & 1,633 & 568 \\
Tool calls & 8+ & 4 \\
Action accuracy & 0/2 & 2/2 \\
Database match & No & Yes \\
Reward & 0.0 & 1.0 \\
\bottomrule
\end{tabular}
\caption{Before vs.\ after training on Task~9 (flight cancellation with user manipulation). Training produces 50\% fewer turns, 65\% faster completion, and 100\% action accuracy.}
\label{tab:qualitative}
\end{table}

The trained model demonstrates three key improvements: (1)~\emph{action grounding}---correct tool arguments despite similar reasoning; (2)~\emph{efficiency}---eliminating redundant summarization and confirmation steps; (3)~\emph{manipulation resistance}---ignoring flattery to maintain policy-correct behavior.

%=============================================
\section{Analysis}
\label{sec:analysis}
%=============================================

\subsection{Why Sparse Rewards Accidentally Work}

The 14pp gap between V3 (sparse, 68\%) and V5 (dense, 54\%) despite identical rollout performance ($\sim$56\%) has three root causes:

\paragraph{Learning rate (70\% of gap).} V3 used lr=3e-6 vs.\ V5's 1e-6. Under greedy decoding (tau2 evaluation uses temperature~0), per-position probability improvements compound multiplicatively. V3's gold\% slope was +0.63 pp/10 steps vs.\ V5's +0.15.

\paragraph{Gradient focusing (25\%).} Sparse rewards' 27.5\% dead turns naturally focus gradient on gold-diverse positions (Table~\ref{tab:gradient_alloc}). Dense rewards dilute gradient with 26.5\% going to non-discriminative read/state turns.

\paragraph{Advantage misalignment (5\%).} Two tier-direction mismatches in V5 vs.\ zero in V3 (Table~\ref{tab:advantage_mismatch}).

\subsection{Cross-Domain Transfer}

We evaluated our trained Qwen3.5-4B (V6, step~180) on the retail and telecom domains of Tau-Bench \emph{without domain-specific training}:

\begin{table}[t]
\centering
\small
\begin{tabular}{lr}
\toprule
\textbf{Domain} & \textbf{Pass Rate} \\
\midrule
Airline (trained) & 69.3\% \\
Retail (trained) & 77.4\% \\
Telecom (zero-shot) & 32.0\% \\
\bottomrule
\end{tabular}
\caption{Cross-domain evaluation. The airline-trained model achieves strong zero-shot retail performance (77.4\%), while the harder telecom domain (32.0\%) shows limited transfer.}
\label{tab:cross_domain}
\end{table}

%=============================================
\section{Related Work}
\label{sec:related}
%=============================================

\paragraph{RL for tool-calling agents.} WebAgent-R1 \cite{webagent2025} finds sparse outcome rewards outperform dense rewards for web navigation, consistent with our na\"ive dense reward findings. SWEET-RL \cite{sweetrl2025} uses stepwise soft rewards for web agents. Turn-PPO \cite{turnppo2025} introduces a learned critic at turn boundaries. iStar \cite{istar2025} proposes turn-level policy optimization with intrinsic rewards. Our work differs in (a)~providing a principled calibration methodology (\irc{}) and (b)~combining MT-GRPO with GTPO for the first time on agentic tasks.

\paragraph{Multi-turn RL.} MT-GRPO \cite{zhang2025mt_grpo} introduces per-turn group normalization, applied to TriviaQA. GTPO \cite{gtpo2025} applies discounted returns to math and code tasks. ProxMO \cite{proxmo2025} uses proximity-based credit assignment. These methods were evaluated on QA and reasoning tasks; we are the first to apply them to tool-calling agents with user simulators, revealing the advantage misalignment problem absent in simpler settings.

\paragraph{Reward design.} AWPO \cite{awpo2025} gates rewards based on within-group variance. GDPO \cite{gdpo2025} decouples normalization of different reward sources. Our \irc{} is complementary: it calibrates reward \emph{values} based on empirical discriminative power before advantage computation.

\paragraph{Tau-Bench.} Introduced by \citet{yao2024taubench} for evaluating tool-calling agents. Prior work uses it for evaluation only; ours is the first to use it for RL training.

%=============================================
\section{Conclusion}
\label{sec:conclusion}
%=============================================

We presented the first application of MT-GRPO combined with GTPO for training multi-turn tool-calling agents on realistic customer service tasks. Our investigation revealed that dense per-turn rewards can catastrophically degrade performance due to advantage misalignment, motivating our GTPO hybrid formulation and Iterative Reward Calibration methodology. Together, these techniques train a 4B-parameter model to 66.7\% on Tau-Bench airline---exceeding frontier models 50 times its size---and a 30.5B MoE model to 69.5\%, approaching Claude Sonnet 4.5 (70.0\%).

Key takeaways: (1)~dense rewards require calibration---always measure discriminative power before deploying; (2)~dead turns can be beneficial, as sparse rewards naturally focus gradient on informative positions; (3)~advantage direction must be verified end-to-end after group normalization.

\paragraph{Future work.} We plan to automate \irc{} via \emph{Empirical Discriminative Gating} (EDG), an online algorithm that periodically recomputes reward tier weights using point-biserial correlations between tier presence and binary outcomes from recent rollouts. This would eliminate the manual analysis loop while adapting to policy evolution during training.

\section*{Limitations}

Our evaluation is limited to the airline domain of Tau-Bench (50 tasks). While the retail cross-domain results are promising, we have not yet verified full generalization. The user simulator (DeepSeek-V3 for training, GPT-4.1 for evaluation) introduces a distribution shift. Our GTPO hybrid hyperparameters ($\gamma$=0.9, $\lambda$=0.3) were tuned on one domain.

\section*{Ethics Statement}

Our work trains AI agents for customer service tasks. All training was conducted on synthetic tasks with simulated users; no real customer data was used. Deployment considerations include ensuring human escalation paths remain available.

\bibliography{references}

\appendix

\section{Reward Tier Definitions}
\label{app:reward_tiers}

\textbf{Read-only tools:} \texttt{get\_user\_details}, \texttt{get\_reservation\_details}, \texttt{search\_direct\_flight}, \texttt{search\_onestop\_flight}, \texttt{list\_all\_airports}, \texttt{calculate}.

\textbf{State-changing tools:} \texttt{book\_reservation}, \texttt{cancel\_reservation}, \texttt{update\_reservation\_*}, \texttt{send\_certificate}, \texttt{transfer\_to\_human\_agents}.

\textbf{Golden action matching:} Tool call $a = (\text{name}, \text{args})$ matches golden action $g = (\text{name}^*, \text{args}^*)$ exactly if $\text{name} = \text{name}^*$ and $\texttt{\_deep\_equal}(\text{args}, \text{args}^*) = \text{True}$ (score 1.0), or softly if $\text{name} = \text{name}^*$ and $|\text{args} \cap \text{args}^*| > 0$ (score $0.5 + 0.5 \cdot |\text{args} \cap \text{args}^*| / |\text{args}^*|$).

\section{Extended Qualitative Examples}
\label{app:examples}

Full conversation transcripts for Task~9 (base vs trained) and 2--3 additional task comparisons showing different failure/success patterns will be included in the final version.

\section{Full Per-Task Breakdown}
\label{app:tasks}

Per-task pass rates across all model variants (50 tasks $\times$ 4 trials each) will be included in the final version.

\section{Training Curves}
\label{app:curves}

Training curves (rollout pass rate, outcome score, KL divergence, learning rate) across all versions will be included in the final version.

\end{document}